\documentclass{article}
\usepackage{spconf,amsmath,graphicx}
\usepackage{booktabs}

\title{Contrastive Learning for Low-light Raw Denoising}
\name{Taoyong Cui\textsuperscript{1}, Yuhan Dong\textsuperscript{1*}}

\address{\textsuperscript{1}Shenzhen International Graduate School, Tsinghua University, Shenzhen, China}
\begin{document}
%
\maketitle
\begin{abstract}
Image/video denoising in low-light scenes is an extremely challenging problem due to limited photon count and high noise. In this paper, we propose a novel approach with contrastive learning to address this issue. Inspired by the success of contrastive learning used in some high-level computer vision tasks, we bring in this idea to the low-level denoising task. In order to achieve this goal, we introduce a new denoising contrastive regularization (DCR) to exploit the information of noisy images and clean images. In the feature space, DCR makes the denoised image closer to the clean image and far away from the noisy image. In addition,  we build a new feature embedding network called Wnet, which is more effective to extract high-frequency information. We conduct the experiments on a real low-light dataset that captures still images taken on a moonless clear night in 0.6 millilux and videos under starlight (no moon present, $<$0.001 lux). The results show that our method can achieve a higher PSNR and better visual quality compared with existing methods.

\end{abstract}
\section{Introduction}
\label{sec:intro}

Due to the very limited photons count and the presence of escapeable noise, it is a great challenge for the quality of low-light photography and videography. In general, we can increase the aperture setting to use high ISO, and extend the exposure time to collect more light. However, high ISOs, while effectively making each pixel more sensitive to light and allowing shorter exposures, can also amplify the noise in each frame. Therefore, more efficient algorithms are needed to obtain high-quality images and videos under low-luminance light.

Over the years, several denoising algorithms have been developed to improve the image/video quality, from classic methods (e.g., spatial domain methods, and transform domain filtering methods) to deep learning based approaches. Each of these methods always attempts to extract the signal from the noise based on some assumptions about the statistical distributions of the image and noise \cite{Monakhova_2022_CVPR}. While these methods have achieved reasonably good performance in some denoising tasks, most are built upon simplistic noise models (Gaussian or Poisson-Gaussian noise), which do not well reflect the severe quantization, bias, and clipping that arise in extreme low-light conditions. Without a good understanding of the structure of the noise in the images/videos, the denoising effect will be poor in the actual low-light denoising task.

The challenge of denoising in low light is well-known in the computational photography community but remains open. Rather than assuming a certain noise model, recently, some learning-based methods automatically account for the low-light noise through a deep neural network and massive training image pairs. However, they always require robust alignment techniques to account for any motion in the scene, which is difficult in the presence of extreme noise. So they always fail to remove such noise in extremely low-light conditions and do not effectively address the color bias. 

In order to tackle this challenge, we propose a new approach with three novel contributions:  \textbf{(1)} We design a new denoising contrastive regularization (DCR) to exploit the information of noisy images and clean images. The DCR makes that the denoised images (anchors) are pulled closer to the clean images (positive samples) and pushed far away from the noisy images (negative samples) in the feature space.
 \textbf{(2)} In DCR, different from the existing methods which adopt a prior model (e.g., pre-trained VGG model) to obtain the feature embedding, we build a novel and task-related one called Wnet, which is more effective to extract high-frequency texture information. \textbf{(3)} Experiments show that our method outperforms several existing methods in terms of quantitative and qualitative results. In addition, we achieve state-of-the-art performance on the starlight dataset. 
\begin{figure}[!tb]
	
	\begin{minipage}[b]{1.0\linewidth}
		\centering
		\centerline{\includegraphics[width=9.5cm]{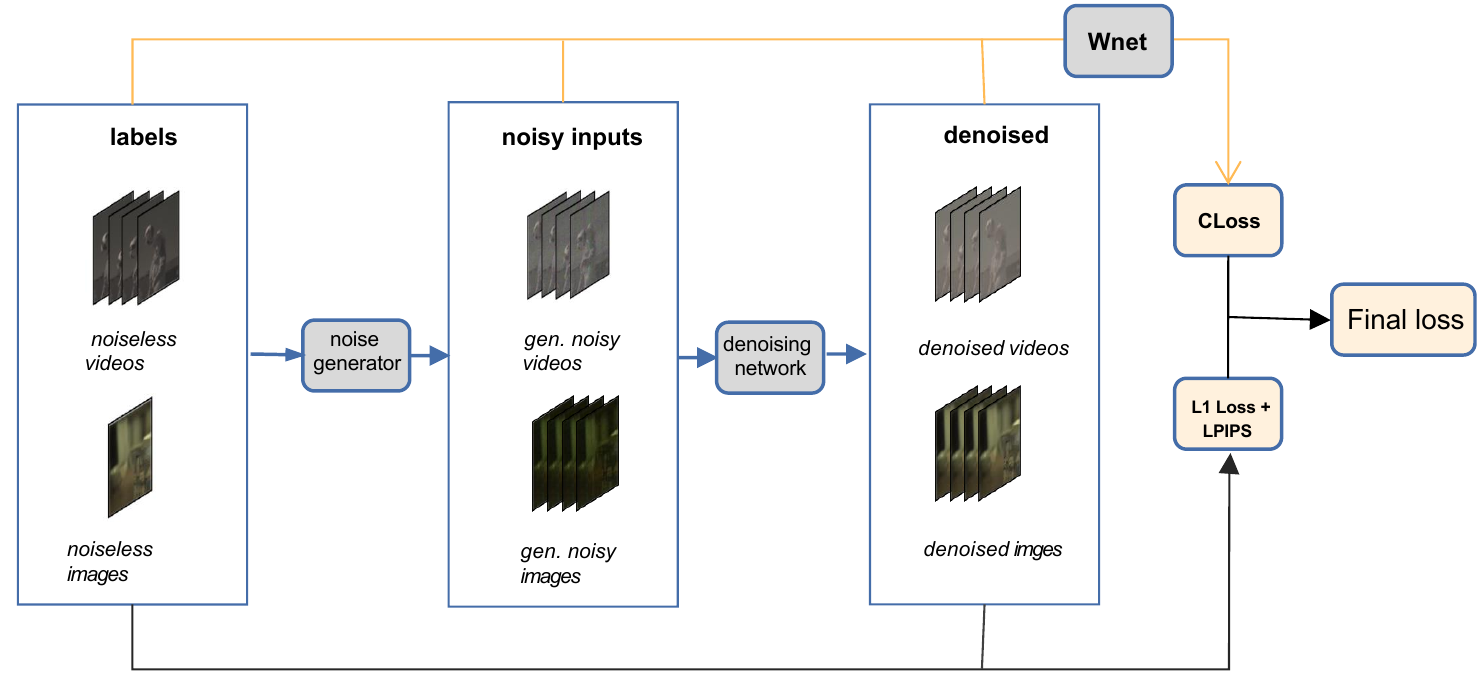}}
		
	\end{minipage}
	\caption{The overall pipeline for low-light denoising is shown in the figure. The inputs to Wnet are noisy frames/images, processed denoised frames/images, and clean frames/images. The output feature encoding of Wnet is used to generate the final Closs, as the arrow points in the figure.
 For denoising network, it takes in 5 noisy
RAW images to produce 1 denoised RAW image.}
	\label{fig:res}
\end{figure}

\section{Related work}
\label{sec:format}

\subsection{Image and Video Denoising} 
Single image/video denoising aims to generate the noise-free image/video from the noisy observation image/video, which can be categorized into prior-based methods and learning-based methods. For prior-based methods \cite{Dong2015,Krull_2019_CVPR,9098612}, the authors attempted to extract the signal from the noise based on certain statistical distribution assumptions, such as Gaussian or Poisson-Gaussian noise. And in \cite{9511233}, the authors systematically studied the noise statistics in the imaging pipeline of CMOS photosensors and formulated a comprehensive noise model that can accurately characterize the real noise structures. In \cite{Monakhova_2022_CVPR}, the authors combined physics-based statistical
methods with GAN-based training techniques to learn to approximate the sensor noise in a data-driven manner. For learning-based methods, these methods \cite{Chen_2019_ICCV,Plotz_2017_CVPR} trained a denoiser using a large amount of clean/noisy image pairs to learn noise information. And for low-light photography, methods to enhance the brightness and contrast of images/videos in dim environments, and perform more global analysis and processing \cite{6012107,7782813,8013257} have been developed. Compared to these methods, we propose a contrastive learning based algorithm to address the challenge of extreme low-light denoising.

\subsection{Contrastive Learning}
When contrastive learning comes to low-level computer vision tasks, the learned global visual representations are inadequate for low-level tasks that call for rich texture and context information \cite{2021arXiv211113924W}. So less work is used for low-level CV tasks. In \cite{Wu_2021_CVPR}, the authors utilized a pre-trained VGG model to obtain the latent embeddings. For the restored image, it takes clean images as positive samples, input hazy images as negative samples, and the restored images as anchor samples. Then contrastive loss is L1 loss, which is conducted with intermediate feature maps extracted from the VGG model. In \cite{2021arXiv211113924W}, the authors proposed a GAN-based perceptual method and a contrastive learning framework for single image super resolution (SISR). However, there are still few works to apply contrastive learning to image/video denoising, as the specialty of this task on exploring a proper and meaningful embedding space. Moreover, different from \cite{Wu_2021_CVPR} and inspired by \cite{2021arXiv211113924W}, we propose a prior-related feature embedding network and a novel pixel-wise contrastive loss for denoising task. 

\section{Our Method}
\label{sec:pagestyle}
To improve the denoised image/video quality, we propose a novel denoising contrastive regularization (DCR), the overall pipeline can be seen in Fig. 1. We first introduce the proposed DCR in Section 3.1 and then present the proposed feature embedding network called Wnet and the pre-training of this network in Section 3.2. At last, we present the proposed contrastive loss in DCR to further improve denoising performance in Section 3.3.

\subsection{Denoising Contrastive Regularization}
Contrastive learning is one of the most powerful approaches for representation learning. It aims at pulling the anchor sample close to the positive samples and pushing it far away from negative samples in latent space \cite{2021arXiv211113924W}. Inspired by contrastive learning \cite{2021arXiv211113924W,Wu_2021_CVPR,2018arXiv180703748V,pmlr-v119-chen20j}, we propose a new contrastive regularization (DCR) to generate higher quality denoised images. Most importantly, in DCR, we need to consider three aspects: first, constructing ``positive" and ``negative" pairs; second, finding the latent feature space of these pairs for comparison; and third, specific loss function. Specifically, in our DCR, the positive pair and negative pair are generated by the group of a clear image $\boldsymbol{P}$ and its denoised image $\boldsymbol{F^{*}}$, and the group of $\boldsymbol{F^{*}}$ and a noisy image $\boldsymbol{N}$, respectively. For the latent feature space, we select the task-related latent space from the proposed Wnet for practical feature embedding. And we propose the Closs function to fully utilize the network.
\subsection{Feature Embedding Network}
As outlined in Section 2, previous work has utilized a pre-trained VGG-based contrastive loss \cite{Wu_2021_CVPR} for low-level computer vision tasks. However, we believe that a task-related embedding network is better because the features obtained by VGG tend to be more focused on high-level semantic information rather than noise information. And we believe that the feature embedding network should not be too deep to avoid learning too much semantic information. Therefore, we develop a simple but efficient network called Wnet to achieve feature embedding. Specifically, in order to learn more noise information, we use the Haar wavelet transform to extract the informative high-frequency components (HL, LH, and HH) and then use CNN to learn these features. For pre-training this feature network, we use the prior noise model \cite{Monakhova_2022_CVPR} to synthesize additional noisy samples and then train with all clean training samples for binary classification, the criterion we choose is Cross-entropy loss. With this simple classification task, Wnet is able to utilize prior knowledge and focus on more noise information in the latent feature space of these “positive” pairs and “negative” pairs for contrast. To enhance the contrastive ability, we extract hidden features from different convolution layers of Wnet as illustrated in Fig. 2.
\begin{figure}[!tbp]
	
	\begin{minipage}[b]{1.0\linewidth}
		\centering
		\centerline{\includegraphics[width=7.5cm]{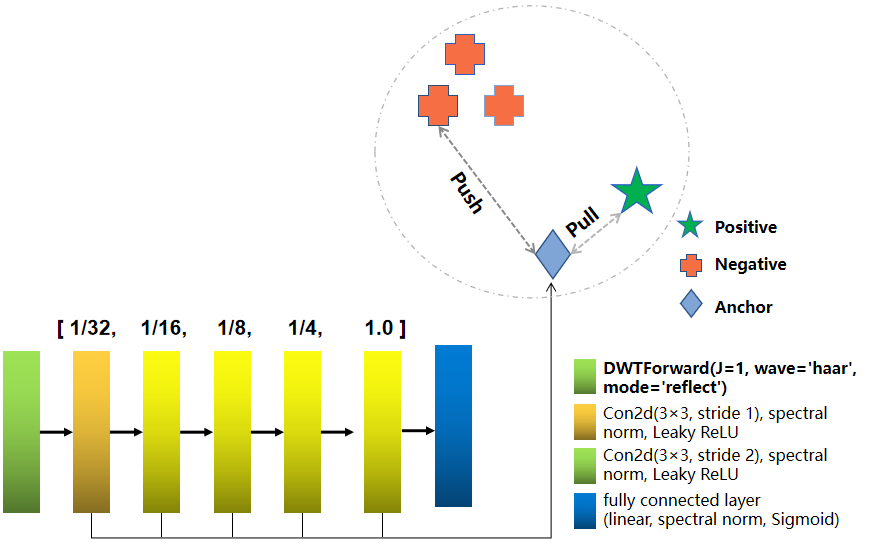}}
		
	\end{minipage}
	\caption{Proposed Wnet.}
	\label{Wnet}
\end{figure}
\subsection{Contrastive Loss}
To fully utilize Wnet, we use multi-intermediate features from Wnet in our contrastive loss and propose a new loss function based on Eq. (1). For a target denoised image $\boldsymbol{F^{*}_{l}}$, its positive and negative counterparts are noted as $\boldsymbol{P_{l}}$ and $\boldsymbol{N_{l}}$ respectively. The feature representations for the denoised image, positive and negative counterparts are noted as $\boldsymbol{f}$, $\boldsymbol{p}$, and $\boldsymbol{n}$, respectively. Our contrastive loss for the $\boldsymbol{l}$-th sample on the $\boldsymbol{i}$-th convolution layer is defined as follows:

\begin{equation}
	\mathcal{L}_{\boldsymbol{Closs}}=\sum_{l=1}^{L} \sum_{i=1}^{N} \omega_{i} \cdot \frac{s\left(G_{i}(p), G_{i}(f)\right)}{s\left(G_{i}(n), G_{i}(f)\right)},
\end{equation}
where $\boldsymbol{G_i}$ = 1, 2, · · ·, N extracts the $\boldsymbol{i}$-th hidden features
from the Wnet, and $w_{i}$ is a weight coefficient. Let’s note the shape of the feature map as $\boldsymbol{C} (channels) \times \boldsymbol{H} (height) \times \boldsymbol{W} (width)$. Inspired by \cite{2021arXiv211113924W, Wu_2021_CVPR}, we adopt the mean value of pixel-wise cosine similarity with L1 loss as the similarity between feature maps. The function $s(f^x,f^y)$ is defined as follows:

\begin{equation}
	s\left(f^{x}, f^{y}\right)=\frac{1}{2 H W} \big(\sum_{h=1}^{H} \sum_{w=1}^{W} \frac{f_{h w}^{x} f_{h w}^{y}}{\left\|f_{h w}^{x}\right\|\left\|f_{h w}^{y}\right\|}+2 |f_{h w}^{x}-f_{h w}^{y}|\big).
\end{equation}
We use a combination of a perceptual loss (LPIPS, which requires a 3-channel image but RAW is 4-channel, so we choose only the first 3 RAW channels for the loss), an L1 loss, and proposed Closs for our training objective, $\boldsymbol{\alpha}$ is a hyperparameter for balancing the main task loss and Closs. And the final loss function is defined as follows:
\begin{equation}
\mathcal{L}_{\boldsymbol{final}}=\mathcal{L}_{\boldsymbol{1}} +\mathcal{LPIPS}+\mathcal{\alpha} \mathcal{L}_{\boldsymbol{Closs}}.
\end{equation}



\section{Experiments}
\label{sec:typestyle}

To test the effectiveness of the DCR, we conducted extensive experiments on a real low-light image/video dataset. All our experiments are implemented by PyTorch 1.8.0 with four NVIDIA V100. The models are trained using the Adam optimizer. The initial learning rate and batchsize are set to 0.0001 and 1, respectively. We set the Closs in Eq. (1) after the latent features of the 1st, 2nd, 3rd, 4th and
5th convolution layers from the pre-trained Wnet, and their corresponding coefficients $w_{i}$, i = 1, · · · , 5 to $\frac{1}{32}$, $\frac{1}{16}$, $\frac{1}{8}$, $\frac{1}{4}$ and $1$, respectively. For datasets, we use all datasets from \cite{Monakhova_2022_CVPR}, including bursts of paired clean/noisy static scenes, 176 clips of clean videos of moving objects (unpaired), 329 video clips from MOT video challenge \cite{leal2015motchallenge} and noisy videos of moving objects in submillilux conditions. All images/videos are captured in RAW format.
To test our DCR performance, the training settings are consistent among ``baseline", ``baseline+vgg+l1", ``baseline+wnet+l1", and ``baseline+DCR" for a fair comparison. The main difference is the choice of feature embedding networks (VGG or Wnet) and specific loss function (L1 loss or Closs) in contrastive regularization. 

\begin{table}[!tbp]
	\newcommand{\tabincell}[2]{\begin{tabular}{@{}#1@{}}#2\end{tabular}}
	\centering
	\caption{Performance on still images from the test set by only training FastDVDnet (Baseline) with still images. ``*" is from \cite{Monakhova_2022_CVPR} including all other compared methods. }
	\resizebox{\linewidth}{!}{
		\begin{tabular}{cccc}
			\toprule
			
			& \tabincell{c}{\textbf{PSNR}} & \tabincell{c}{\textbf{SSIM}} &  \tabincell{c}{\textbf{LPIPS}}
			\\
			\midrule
			\midrule
			\tabincell{c}Noise2Self \cite{pmlr-v97-batson19a}* & 20.11 & 0.210  & 0.545 \\
			\midrule
			\tabincell{c}Unprocesing \cite{Brooks_2019_CVPR}* & 12.86 & 0.249  & 0.355 \\
			\midrule
			\tabincell{c}L2SID \cite{Chen_2018_CVPR}* & 26.90 & \textbf{0.892}  & 0.198 \\
			\midrule
			\tabincell{c}Baseline (FastDVDnet) \cite{Tassano_2020_CVPR} & 30.45 & 0.841  & 0.076 \\
			\midrule
			\tabincell{c}Baseline+(VGG+L1) \cite{Wu_2021_CVPR}   & 29.06 & 0.823  & 0.074  \\
			\midrule
			\tabincell{c}Baseline+(Wnet+L1)    & 31.94 & 0.854  & 0.070  \\
			\midrule
			\tabincell{c}{\textbf{Baseline+DCR (Ours)}}    & \textbf{31.97} & 0.870 & \textbf{0.065} \\
			\bottomrule
			
		\end{tabular}
	}
	
	\label{stillimages}
\end{table}

\begin{figure}[!tb]
	
	\begin{minipage}[b]{1.0\linewidth}
		\centering
		\centerline{\includegraphics[width=8.5cm]{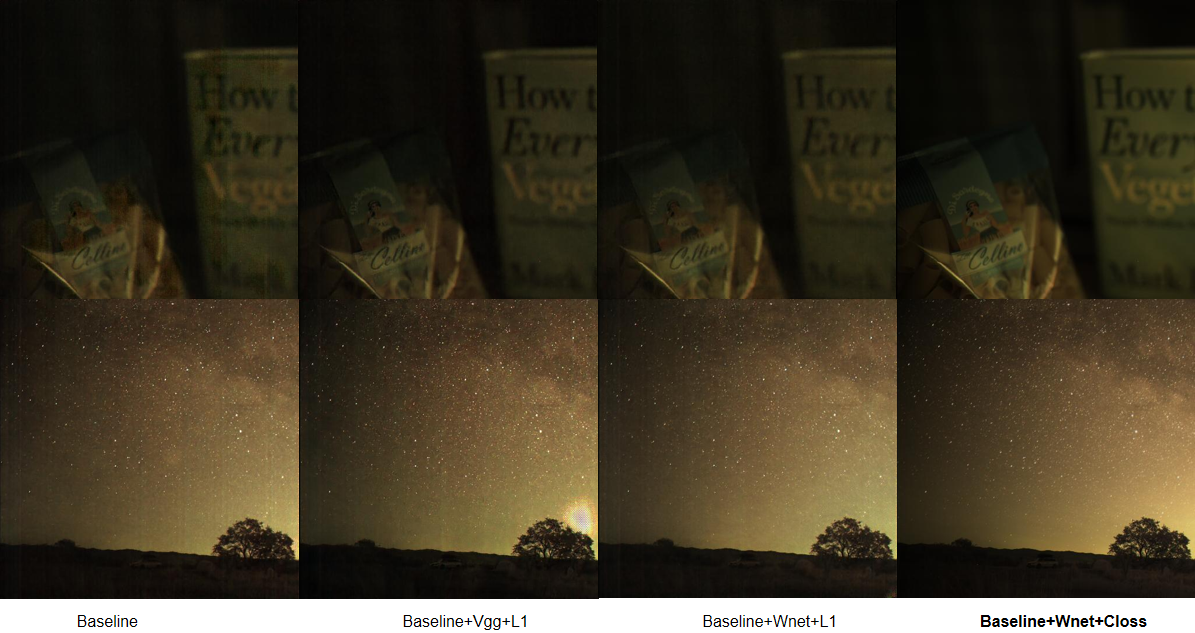}}
		
	\end{minipage}
	\caption{Example of performance on still images from the test set by only training FastDVDnet with still images. }
	\label{still}
\end{figure}
\subsection{Experiment on Still Images}
\label{sec:majhead}
For denoising still images, we build the FastDVDnet \cite{Monakhova_2022_CVPR} model, which takes in 5
noisy RAW images to produce 1 denoised RAW image. 80\% still images (synthetic noisy samples/clean samples) to train the denoising image model, and 20\% still images (real samples) to test performance. We feed in 5 noisy clips to each denoiser, then compare against a single still ground-truth image. These results are summarized in Table 1, with image examples shown in Fig. 3. Although the improvement achieved with Closs relative to the ``baseline+Wnet+l1" approach may not be immediately apparent based on PSNR, the visual quality is significantly improved, as shown in Fig. 3.

\subsection{FastDVDnet on Full Datasets}

Both the paired still images and the clean videos of moving objects are used to train the denoiser. We build the FastDVDnet model to evaluate our proposed DCR performance. In this experiment, we only change the dataset without changing the denoising model. By this way, we test our DCR generalization in different scenarios (including still and moving). Furthermore, Fig. 4 shows a comparison of restored quality on noisy samples from the test set, wherein the proposed DCR can help the denoiser model to improve the restored quality.

\begin{table}[!tbp]
	\newcommand{\tabincell}[2]{\begin{tabular}{@{}#1@{}}#2\end{tabular}}
	\centering
	\caption{Performance on still images from the test set by training FastDVDnet with still images and videos.}
	\resizebox{\linewidth}{!}{
		\begin{tabular}{cccc}
			\toprule
			
			& \tabincell{c}{\textbf{PSNR}} & \tabincell{c}{\textbf{SSIM}} &  \tabincell{c}{\textbf{LPIPS}}
			\\
			\midrule
			\midrule
			\tabincell{c}Baseline & 31.44 & 0.852  & 0.0696 \\
			\midrule
			\tabincell{c}Baseline+(VGG+L1)\cite{Wu_2021_CVPR}   & 30.23 & 0.842  & 0.0635  \\
			\midrule
			\tabincell{c}{\textbf{Baseline+DCR (Ours)}}    & \textbf{32.23} & \textbf{0.853} & \textbf{0.0630} \\
			\bottomrule
			
		\end{tabular}
	}
	
	\label{video1}
\end{table}
\begin{figure}[!tbp]
	
	\begin{minipage}[b]{1.0\linewidth}
		\centering
		\centerline{\includegraphics[width=8.5cm]{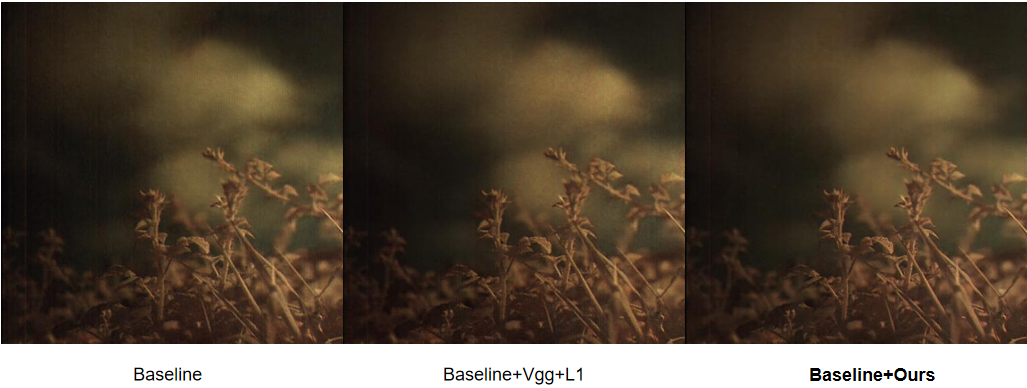}}
		
	\end{minipage}
	\caption{Example of performance by training with still images and moving videos. }
	\label{still+}
\end{figure}
\subsection{Modified FastDVDnet on Datasets}
In this experiment, we compare the performance of the DCR-based model with state-of-the-art methods for starlight video denoising. The other two experiments in Section 4.1 and Section 4.2 mainly focus on evaluating the performance of DCR on low-light images. We train the denoiser (Modified FastDVDnet \cite{Mildenhall_2018_CVPR}, replacing the U-Net denoising blocks with an HRNet, which
leads to better temporal consistency) on a combination of synthetic noisy video clips and real still images. First, we pre-train our network for 500 epochs using a combination of real paired stills, synthetic noisy clips, and synthetic noisy clips from the MOT dataset to help prevent overfitting. After pretraining, we refine the model on real still images and synthetic clips. All images are cropped to 256×256 patches throughout training. Next, we qualitatively compare our performance on the unlabeled dataset of submillilux video clips, and the results are shown in Figure 5.
\begin{table}[!tbp]
	\newcommand{\tabincell}[2]{\begin{tabular}{@{}#1@{}}#2\end{tabular}}
	\centering
	\caption{Performance on still images from the test set by training Modified FastDVDnet with still images and videos. ``retrained" represents that we retrain the model according to the method of the paper \cite{Monakhova_2022_CVPR}, ``*" data is from \cite{Monakhova_2022_CVPR}.}
	\resizebox{\linewidth}{!}{
		\begin{tabular}{cccc}
			\toprule
			
			& \tabincell{c}{\textbf{PSNR}} & \tabincell{c}{\textbf{SSIM}} &  \tabincell{c}{\textbf{LPIPS}}
			\\
			\midrule
			\tabincell{c}FastDVDnet* & 23.8 & 0.618  & 0.282 \\
			\midrule

			\tabincell{c}Baseline (retrained) & 27.3 & 0.837  & 0.091 \\
			\midrule
			\tabincell{c}Baseline (\textbf{SOTA})* & 27.7 & \textbf{0.931}  & 0.078 \\
			\midrule
			\tabincell{c}Baseline+(VGG+L1)   & 28.1 & 0.831  & 0.103  \\
			\midrule
			\tabincell{c}{\textbf{Baseline+DCR (Ours)}}    & \textbf{30.5} & 0.889 & \textbf{0.060} \\
			\bottomrule
			
		\end{tabular}
	}

	\label{vedio2}
\end{table}
\begin{figure}[t!b]
	
	\begin{minipage}[b]{1.0\linewidth}
		\centering
		\centerline{\includegraphics[width=8.5cm]{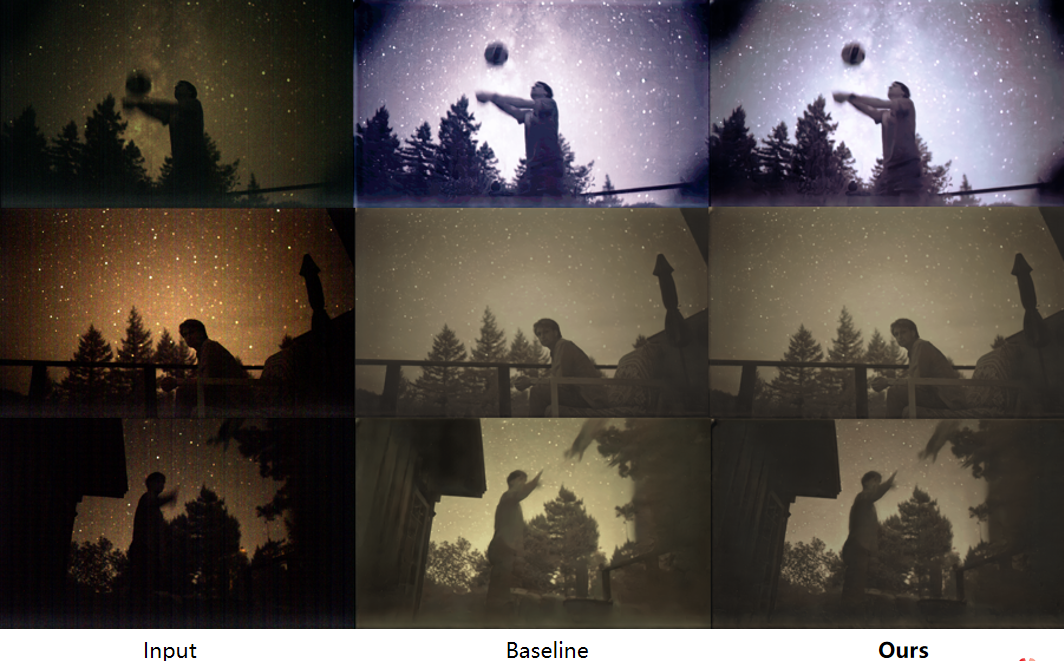}}
		
	\end{minipage}
	\caption{Visual comparison of the Modified FastDVDnet performance on unlabeled submillilux videos (after the same post-processing). We compare it with the current SOTA in this dataset, obviously, DCR can help existing benchmarks achieve better performance.}
	\label{submil}
\end{figure}
\section{Conclusion}
\label{sec:print}

In this work, we proposed a novel contrastive
regularization (DCR) to further improve the performance of existing low-light image/video denoising approaches. For the feature embedding task, we develop a task-related Wnet to better extract high-frequency noise information. At last, extensive experiments have shown that our proposed approach can help existing benchmarks achieve better performance.

\bibliographystyle{IEEEbib}
\bibliography{strings,refs}

\end{document}